\documentclass[conference]{IEEEtran}
\IEEEoverridecommandlockouts
\usepackage{cite}
\usepackage{amsmath,amssymb,amsfonts}
\usepackage{algorithmic}
\usepackage{graphicx}
\usepackage{textcomp}
\usepackage{xcolor}
\usepackage{booktabs}
\usepackage[T1]{fontenc}
\usepackage{glossaries-extra}
\usepackage{float}
\usepackage{xcolor}
\usepackage{url}
\usepackage[misc]{ifsym}
\usepackage{graphicx}
\def\BibTeX{{\rm B\kern-.05em{\sc i\kern-.025em b}\kern-.08em
    T\kern-.1667em\lower.7ex\hbox{E}\kern-.125emX}}

\usepackage{scalerel}
\usepackage{tikz}
\usetikzlibrary{svg.path}

\definecolor{orcidlogocol}{HTML}{A6CE39}
\tikzset{
  orcidlogo/.pic={
    \fill[orcidlogocol] svg{M256,128c0,70.7-57.3,128-128,128C57.3,256,0,198.7,0,128C0,57.3,57.3,0,128,0C198.7,0,256,57.3,256,128z};
    \fill[white] svg{M86.3,186.2H70.9V79.1h15.4v48.4V186.2z}
                 svg{M108.9,79.1h41.6c39.6,0,57,28.3,57,53.6c0,27.5-21.5,53.6-56.8,53.6h-41.8V79.1z M124.3,172.4h24.5c34.9,0,42.9-26.5,42.9-39.7c0-21.5-13.7-39.7-43.7-39.7h-23.7V172.4z}
                 svg{M88.7,56.8c0,5.5-4.5,10.1-10.1,10.1c-5.6,0-10.1-4.6-10.1-10.1c0-5.6,4.5-10.1,10.1-10.1C84.2,46.7,88.7,51.3,88.7,56.8z};
  }
}

\newcommand\orcidicon[1]{\href{https://orcid.org/#1}{\mbox{\scalerel*{
\begin{tikzpicture}[yscale=-1,transform shape]
\pic{orcidlogo};
\end{tikzpicture}
}{|}}}}

\usepackage{hyperref} 

\begin{document}

\title{Large Language Model in Medical Informatics: Direct Classification and Enhanced Text Representations for Automatic ICD Coding\\

}

\author{
\IEEEauthorblockN{
Zeyd Boukhers\IEEEauthorrefmark{1}\IEEEauthorrefmark{4}\IEEEauthorrefmark{3}\orcidicon{0000-0001-9778-9164}\Letter,
AmeerAli Khan\IEEEauthorrefmark{1}\orcidicon{0009-0001-0020-1262},
Qusai Ramadan\IEEEauthorrefmark{3}\IEEEauthorrefmark{5} \orcidicon{0000-0001-8159-918X},
Cong Yang\IEEEauthorrefmark{2} \orcidicon{0000-0002-8314-0935}
}

\IEEEauthorblockA{\IEEEauthorrefmark{1}Fraunhofer Institute for Applied Information Technology FIT, Germany \\
Emails: zeyd.boukhers@fit.fraunhofer.de, ameerali.khan@fit.fraunhofer.de}
\IEEEauthorblockA{\IEEEauthorrefmark{4}University Hospital of Cologne, Germany}
\IEEEauthorblockA{\IEEEauthorrefmark{3}University of Koblenz, Germany \\
Email: qramadan@uni-koblenz.de}
\IEEEauthorblockA{\IEEEauthorrefmark{5}University of Southern Denmark, Denmark}
\IEEEauthorblockA{\IEEEauthorrefmark{2}Soochow University, China \\
Email: cong.yang@suda.edu.cn}
}

\maketitle

\begin{abstract}
Addressing the complexity of accurately classifying International Classification of Diseases (ICD) codes from medical discharge summaries is challenging due to the intricate nature of medical documentation. This paper explores the use of Large Language Models (LLM), specifically the LLAMA architecture, to enhance ICD code classification through two methodologies: direct application as a classifier and as a generator of enriched text representations within a Multi-Filter Residual Convolutional Neural Network (MultiResCNN) framework. We evaluate these methods by comparing them against state-of-the-art approaches, revealing LLAMA's potential to significantly improve classification outcomes by providing deep contextual insights into medical texts.
\end{abstract}


\section{Introduction}

In the past decade, advancements in Deep Learning (DL) and Natural Language Processing (NLP) have revolutionized healthcare research, propelled by the growth in health data~\cite{nguyen2016mathtt}. These advancements have successfully improved the interpretation of medical images and the processing of Electronic Health Records (EHR) through Deep Neural Networks~\cite{li2020icd}. Such technologies are crucial in automatically assigning International Classification of Diseases (ICD) codes, globally vital for healthcare documentation and administration~\cite{nadathur2010maximising}.

Research on automating ICD coding, which began over two decades ago, has shifted from manual feature creation to employing advanced machine learning techniques~\cite{li2020icd}.

Despite these advancements, mapping unstructured medical texts to specific ICD codes remains challenging, mainly due to the complexity of medical language. Current models often fail to capture the full contextual and semantic depth required for accurate ICD coding. Recently, Generative Large Language Models (LLMs) like LLAMA and Mixtral have begun to improve standards of accuracy and efficiency in text analysis. These models use advanced linguistic understanding to enhance the correlation between vast documents and entities~\cite{chang2023survey}.

This paper explores the use of Large Language Models in automatic ICD coding, particularly employing LLAMA-2. These models are employed for direct classification and for generating enriched text representations for processing by a CNN-based classifier, namely a Multi-Filter Residual Convolutional Neural Network (MultiResCNN). These applications aim to fully exploit the semantic capabilities of LLAMA-2 (7b) to enhance its effectiveness in medical text interpretation and ICD code classification.

Our main contributions are:
\begin{itemize}
    \item Adapting LLAMA-2 (7b) for direct ICD code classification and evaluating its effectiveness beyond typical generative uses.
    \item Using LLAMA-2 (7b) to generate enriched text representations, processed by MultiResCNN for enhanced classification.
    \item Rigorously evaluating these methods on the MIMIC-III dataset and comparing their performances with established baselines.
\end{itemize}

\section{Related Work}
\label{sec:rel}

This section reviews relevant ICD coding techniques divided into three categories.

\subsection{Traditional Machine Learning Techniques}
ICD coding has relied on traditional machine learning methods, using manually crafted features and established algorithms~\cite{scheurwegs2016data}. Support vector machines (SVM) have been prevalent and applied across diverse healthcare settings for classifying ICD codes \cite{koopman2015automatic}.

Other approaches, including regular expression-based mapping and adaptive data processing, have enhanced accuracy and efficiency~\cite{zhou2020construction}. Feature engineering, particularly using gradient boosting for large datasets of discharge texts, has also been significant \cite{diao2021automated}.

\subsection{Neural Network and Knowledge-Enhanced Techniques for ICD Coding}

Neural networks have revolutionized ICD coding with their flexibility and effectiveness. Techniques such as character-level LSTMs, GRUs, and Hierarchical Attention-bidirectional GRUs (HA-GRUs) have refined text-to-ICD code alignment \cite{baumel2018multi}. Additionally, innovative mixed embedding models optimize word and label vector predictions \cite{wang2018joint}, while ensemble strategies merge CNNs, LSTMs, and decision trees to boost accuracy \cite{xu2019multimodal}. The CAML and Multi-Filter Residual Convolutional Neural Network (MultiResCNN) models employ advanced mechanisms like label attention and multi-filter networks to enhance interpretability and coding performance on MIMIC datasets \cite{li2020icd}.

Moreover, integrating external knowledge has significantly enhanced ICD code predictions. This includes refining term embeddings with medical definitions, enriching the CAML model with Wikipedia data for rare diseases \cite{bai2019improving}, and utilizing medical ontologies and ICD descriptions to improve predictions \cite{bao2021medical}. Our earlier work \textbf{KG-MultiResCNN} has outperformed baselines by effectively leveraging structured external knowledge to capture complex medical relationships \cite{boukhers2023knowledge}.

\subsection{LLM-based Approaches}

The use of LLMs for ICD coding represents a recent focus in medical informatics. Inspired by the transformative effect of pretrained Transformer models, their application in automated ICD coding has attracted significant interest. Silvestri et al.~\cite{9219640} emphasize the potential of transformers to enhance ICD-10 coding accuracy across languages through multilingual capabilities.

Biswas et al.~\cite{10.1007/978-3-030-77211-6_56} introduced TransICD, a transformer-based model that employs a code-wise attention mechanism to capture token interdependencies within documents, addressing code frequency imbalances with a label distribution aware margin (LDAM) loss function. Liu et al. \cite{Liu_2022} developed XR-LAT, a Transformer-based approach using a recursively trained model chain on a hierarchical code tree, enhanced by label-wise attention and knowledge transfer, significantly improving macro-AUC on the MIMIC-III and MIMIC-II datasets.

\section{LLAMA for ICD Coding}
\label{sec:app}

This work utilizes LLAMA-2, a Large Language Model, for ICD code classification ~\cite{touvron2023llama}. We adapt LLAMA-2 as a classifier and as text representations for another classifier, focusing on the mathematical foundations and strategic implementations that address the challenges of medical text analysis.

\subsection{LLM as Classifier}

The $llm\_classifier$ configuration employs a fine-tuning strategy on the LLM to function as a robust sequence classifier. This setup incorporates a specialized classification head for navigating the discrete ICD code space, ranging from 50 top codes to 8239 full codes. The classification process is defined by the equation:


\begin{equation}
    P(y|x; \theta) = \text{softmax}(W_h h(x; \theta) + b)
\end{equation}

\noindent where $x$ represents the input text from discharge summaries, and $y$ is the predicted ICD code. $\theta$ denotes the LLAMA model parameters, $h(x; \theta)$ the final hidden representation, and $W_h$ and $b$ are the weights and bias of the classification layer. This setup effectively captures the complex semantics of medical texts.

The optimization objective in the $llama\_classifier$ configuration is to minimize cross-entropy loss between the predicted and actual ICD codes, defined as:

\begin{equation}
L(\theta) = -\sum_{i=1}^{N} \sum_{c=1}^{C} y_{i,c} \log P(y=c|x_i; \theta),
\end{equation}

\noindent where $y_{i,c}$ is the ground truth label, and $P(y=c|x_i; \theta)$ is the predicted probability of class $c$ for input $x_i$, parameterized by $\theta$.

\subsection{LLAMA as text representation}

\begin{figure}
    \centering
    \includegraphics[width=1\linewidth]{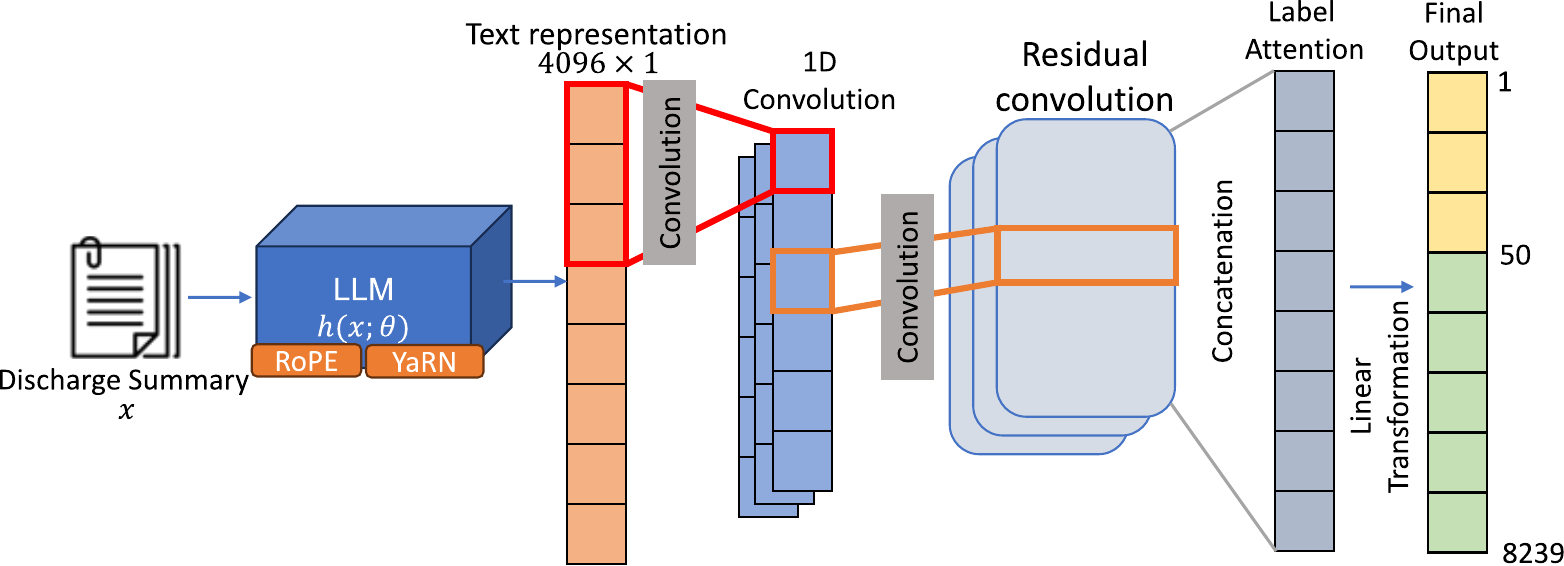}
    \caption{Architecture of the LLAMA-2 Model for Text Representation using RoPE and YaRN techniques. The process culminates in a label attention mechanism that outputs predictions across ICD classes.}
    \label{fig:ovr2}
\end{figure}

Beyond direct classification, LLAMA architecture is used to generate enriched text representations for a deeper semantic understanding of medical texts. Discharge summaries are transformed into high-dimensional vectors capturing essential clinical information, using the raw LLAMA-2 model without further fine-tuning. The goal is to assess the informativeness of these embeddings about their clinical content. This is quantified as the discrepancy between the embeddings and their idealized representations, which would yield perfect classification accuracy with any classifier or architecture $\gamma$. This process is depicted in Fig~\ref{fig:ovr2}. Details on the MultiResCNN architecture are provided in \cite{li2020icd,boukhers2023knowledge}.

\paragraph{Generating Text Representations:} To address the length of clinical texts, the model’s context window is expanded beyond the standard $4096$ tokens using the Rotatory Position Embedding (RoPE) technique \cite{su2023roformer} and further refined by the 'Yet another RoPE extension method' (YaRN) \cite{peng2023yarn}. These techniques adaptively adjust the context size, enhancing the model’s handling of extensive texts. The RoPE technique rotates positional encodings based on sequence positions, while YaRN dynamically scales these rotations according to document length, as modelled by: $S_{new} = S_{original} \cdot \alpha$. where $S_{new}$ is the expanded context size, $S_{original}$ the initial size, and $\alpha$ the scaling factor. The positional encoding function is adjusted as:

\begin{equation}
    PE_{pos} = \sin(pos \cdot \beta), \cos(pos \cdot \beta)
\end{equation}

\noindent with $\beta = \log(\alpha) \cdot \rho$, where $\beta$ modifies the embedding based on position and $\rho$ is an adjustment coefficient. These modifications ensure the model dynamically adjusts to larger textual inputs, improving comprehension and classification accuracy for extensive medical documents.

\paragraph{MultiResCNN Classifier}

Following the work of Li and Yu \cite{li2020icd}, we developed a multi-filter 1-dimensional MultiResCNN to map clinical text representations to ICD codes. Given the computational challenges of managing a high-dimensional space of $4096$ dimensions, we employ two strategies for dimensionality reduction: pre-processing with CNN layers and integrating multiple residual layers within MultiResCNN itself. Both strategies collectively reduce computational load and enhance classification performance.

The initial dimension reduction is handled by:
\begin{equation}
    V_{reduced} = \text{Conv}(V_{4096}; \theta_{conv})
\end{equation}

Here, $V_{4096}$ is the dense text representation, and $\text{Conv}(\cdot)$ represents the convolution operation parameterized by $\theta_{conv}$, resulting in a dimensionally optimized vector $V_{reduced}$.

Further dimensionality reduction within MultiResCNN is achieved through a multi-layer residual mechanism that compresses the data to $128$ dimensions:
\begin{equation}
V_{final} = \sum_{k=1}^{K} \text{ReLU}(\text{ResLayer}_k(V_{reduced}; \theta_{k})),
\end{equation}
\noindent where $ResLayer_k(.;\theta_k)$ represents the $k$-th residual layer operation, and $\text{ReLU}(.)$ is the activation function, allowing the model to handle complex data patterns efficiently.

\section{Experimental Results}
\label{sec:res}
In this section, we evaluate the effectiveness of both applications of LLAMA against baseline approaches. The codes of all the experiments can be found in the GitLab repository\footnote{\url{https://gitlab.com/fdda1/automatic-diagnosis-from-clinical-texts/-/tree/main/icd_coding}}

\paragraph{Dataset:} This paper uses the MIMIC-III database~\cite{johnson2016mimic}, which includes de-identified "Discharge Summaries" for around 40,000 patients. Our study, following references~\cite{boukhers2023knowledge, li2020icd}, focuses on the full 4216 ICD codes and the top 50 codes.

\paragraph{Evaluation Metrics:}
To evaluate ICD code assignment, reducing false negatives is prioritized to avoid missing critical diagnoses. The key metrics include Area Under the Receiver Operating Characteristic Curve (AUC), F1 Score, and Precision at thresholds P@5, P@8, and P@15. We use both micro and macro averaging to address class imbalance and provide representative scores across different ICD codes.

\paragraph{Baselines:}

We compare our approach with established models, including MultiResCNN~\cite{li2020icd} and several benchmark approaches: \textbf{C-MemNN}~\cite{prakash2017condensed}, \textbf{C-LSTM-Att}~\cite{shi2017towards}, \textbf{CAML}~\cite{mullenbach2018explainable}, \textbf{DR-CAML}~\cite{mullenbach2018explainable}, \textbf{KG-MultiResCNN}~\cite{boukhers2023knowledge}, \textbf{XR-LAT-BootstrapHyperC}~\cite{LIU2023102662,Liu_2022}, \textbf{TransICD}~\cite{10.1007/978-3-030-77211-6_56}.

All comparisons adhere to the experimental protocol established by MultiResCNN~\cite{li2020icd}. Many studies were excluded from comparison due to incompatible methodologies or unverifiable results.

 \paragraph{Results:}
 
Our analysis starts with the top 50 ICD codes to fine-tune the LLAMA-based classifiers. After optimizing settings, we expanded to the full-code dataset. Hyperparameters for LLAMA-2. The hyperparameters for the MultiResCNN were also empirically optimized.

As mentioned in Section~\ref{sec:app}, the dimensionality of text representation is managed via CNN Reduction and Residual Layer strategies. We tested CNN layers with dimensions $1024$, $768$, $512$, $300$, and $100$, referred to as \textbf{CNN-1024}, \textbf{CNN-768}, \textbf{CNN-512}, \textbf{CNN-300}, and \textbf{CNN-100} respectively. The \textbf{CNN-1024\_map\_256} model, reducing from $4096$ to $1024$ and then to $256$ dimensions, emerged as the most efficient, balancing computational demands and performance.

For residual mechanisms, we experimented with configurations \textbf{Residual-4096} and \textbf{Residual-128}, the latter compressing to $128$ dimensions. Notably, \textbf{Residual-128} slightly outperforms \textbf{CNN-1024\_map\_256}, but the latter maintains optimal performance with reduced computational load.

We employed the \textbf{CNN-1024\_map\_256} dimension reduction strategy for training \textbf{LLAMA2-R+MRCNN} configuration. Results, detailed in Table~\ref{tab:combined_llama_results}, show these models outperforming baselines such as \textbf{MultiResCNN} and \textbf{DR-CAML}, particularly in metrics like F1 score and AUC, highlighting their effectiveness in handling clinical texts.

For full code classification, \textbf{LLAMA2-C} underperformed, likely due to limited training data and the LLAMA architecture's unsuitability for the granularity of ICD codes. In contrast, \textbf{LLAMA2-R+MRCNN} performed comparably to baselines but did not exceed the performance seen with 50 codes. \textbf{KG-MultiResCNN} notably surpassed these models, suggesting the benefits of integrating external knowledge to address data sparsity and label granularity.

The \textbf{XR-LAT-BHC} model excelled without external knowledge, possibly due to its use of hyperbolic embeddings, which are effective for capturing hierarchical relationships. However, its high computational demand could limit practical application.

\begin{table*}[ht!]
  \centering
  \footnotesize
  \begin{tabular}{p{3cm}|p{1cm}p{1cm}p{1cm}p{1cm}p{1cm}|p{1cm}p{1cm}p{1cm}p{1cm}p{1cm}p{1cm}}
    \toprule
    & \multicolumn{5}{c|}{Top 50 Codes} & \multicolumn{6}{c}{Full Codes} \\
    \midrule
    & F1 Macro & F1 Micro & AUC Macro & AUC Micro & P@5 & F1 Macro & F1 Micro & AUC Macro & AUC Micro & P@8 & P@15 \\
    \midrule
    C-MemNN \cite{prakash2017condensed} & - & -  & 0.833 & - & 0.420 & - & - & - & - & - & -\\
    C-LSTM-Att \cite{shi2017towards} & - & 0.532 & - & 0.900  & - & - & - & - & - & - & -\\
    TransICD~\cite{10.1007/978-3-030-77211-6_56                    } & 0.562 & 0.644 & 0.894 & 0.923 & 0.617  & - & - & - & - & - & -\\
    MultiResCNN~\cite{li2020icd} & 0.606 & 0.670 & 0.899 & 0.928 & 0.641 & 0.085 & 0.552 & 0.910 & 0.986 & 0.734 & 0.584 \\
    KG-MultiResCNN$^*$~\cite{boukhers2023knowledge} & \textbf{0.645} & 0.691 & - & - & - & 0.102 & \textbf{0.651} & - & - & - & - \\
    DR-CAML$^*$ \cite{mullenbach2018explainable} & 0.576 & 0.633 & 0.884 & 0.916 & 0.618 & 0.086 & 0.529 & 0.897 & 0.985 & 0.690 & 0.548 \\
    CAML \cite{mullenbach2018explainable} & 0.532 & 0.614 & 0.875 & 0.909 & 0.609 & 0.088 & 0.539 & 0.895 & 0.986 & 0.709 & 0.561 \\
    XR-LAT-BHC~\cite{LIU2023102662,Liu_2022} & - & - & - & - & - & \underline{\textbf{0.108}} & \underline{0.583} & \underline{\textbf{0.946}} & \underline{\textbf{0.99}} & \underline{\textbf{0.749}} & \underline{\textbf{0.599}}\\
    \textbf{LLAMA2-C}& 0.5802 & 0.6357 & 0.9011 & 0.9241 & 0.6261 & 0.0241 & 0.3909 & 0.8604 & 0.9793 & 0.6304 & 0.4789 \\
    \textbf{LLAMA2-R+MRCNN} & 0.6258 & \textbf{0.6912} & \textbf{0.9138} & \textbf{0.9361} & \textbf{0.6517} & 0.0688 & 0.5324 & 0.8937 & 0.9838 & 0.7364 & 0.5811 \\
    \bottomrule
  \end{tabular}
  \caption{Comparative evaluation results for the top-50 codes and full codes. An asterisk $^*$ next to a method denotes the incorporation of external knowledge. \textbf{Bold} results highlight the best performance, while \underline{underlined} results indicate the top performance among methods that do not utilize external knowledge.}
  \label{tab:combined_llama_results}
\end{table*}

\section{Conclusion}
\label{sec:conc}

In this paper, we explored the application of the LLAMA-2 model for ICD code classification, utilizing it both as a direct classifier and as a generator of enhanced text representations. The results show that while the standalone LLAMA-2 model underperforms, especially with a broad set of ICD codes, their combination with Multi-ResCNN substantially improves performance for a smaller code set by leveraging rich semantic information. However, for larger code sets, performance matches baseline models, highlighting challenges related to label granularity and training data sparsity.

Future work will focus on integrating external knowledge, such as knowledge graphs, to address training data sparsity and enhance model performance. We also propose developing a unified architecture that combines large language models with residual CNNs, aiming to improve both training efficiency and model accuracy.

\section*{Acknowledgement}

This work is supported by the Fraunhofer Cluster of Excellence Cognitive Internet Technologies through the \textbf{CARE-LLM} project, and the LLM Internal Initiative of Fraunhofer FIT, through the \textbf{ELMTEX} project.

%
%
%
\bibliographystyle{ieeetr}
\bibliography{references}

\end{document}